\documentclass[11pt]{article}
\usepackage[preprint]{acl}
\usepackage{times}
\usepackage{latexsym}
\usepackage[T1]{fontenc}
\usepackage[utf8]{inputenc}
\usepackage{microtype}
\usepackage{inconsolata}
\usepackage{graphicx}
\usepackage{amsmath}
\usepackage{booktabs}
\usepackage{etoolbox}
\newif\ifPZeroResultsAvailable\PZeroResultsAvailabletrue
\newcommand{\PZeroEvalN}{100}
\newcommand{\PZeroCap}{2048}
\newcommand{\PZeroPrimarySeed}{1357911}
\newcommand{\PZeroQwenFullCorrect}{83}

\newcommand{\PZeroQwenFirstboxCorrect}{82}
\newcommand{\PZeroQwenFirstboxNull}{14}

\newcommand{\PZeroQwenSolutionCorrect}{90}

\newcommand{\PZeroQwenSolutionActivated}{16}

\newcommand{\PZeroQwenSolutionInjected}{144}

\newcommand{\PZeroQwenWholeSeconds}{14149.60}
\newcommand{\PZeroQwenFirstVsFullSlotSaving}{1.77}

\newcommand{\PZeroQwenSolVsFullSlotSaving}{2.38}

\newcommand{\PZeroQwenGains}{8}
\newcommand{\PZeroQwenHarms}{0}
\newcommand{\PZeroQwenAccuracyDelta}{8.0}
\newcommand{\PZeroQwenAccuracyLow}{3.0}
\newcommand{\PZeroQwenAccuracyHigh}{14.0}
\newcommand{\PZeroQwenSlotSaving}{0.628}
\newcommand{\PZeroQwenSlotLow}{0.371}
\newcommand{\PZeroQwenSlotHigh}{0.886}
\newcommand{\PZeroQwenWallSaving}{0.800}

\newcommand{\PZeroQwenHarmUCB}{3.62}
\newcommand{\PZeroRoneFullCorrect}{80}

\newcommand{\PZeroRoneFirstboxNull}{12}

\newcommand{\PZeroRoneSolutionCorrect}{87}

\newcommand{\PZeroRoneSolutionActivated}{12}

\newcommand{\PZeroRoneSolutionInjected}{108}

\newcommand{\PZeroRoneWholeSeconds}{9934.77}
\newcommand{\PZeroRoneFirstVsFullSlotSaving}{13.93}

\newcommand{\PZeroRoneSolVsFullSlotSaving}{14.44}

\newcommand{\PZeroRoneGains}{7}
\newcommand{\PZeroRoneHarms}{0}
\newcommand{\PZeroRoneAccuracyDelta}{7.0}
\newcommand{\PZeroRoneAccuracyLow}{2.0}
\newcommand{\PZeroRoneAccuracyHigh}{12.0}
\newcommand{\PZeroRoneSlotSaving}{0.602}
\newcommand{\PZeroRoneSlotLow}{0.323}
\newcommand{\PZeroRoneSlotHigh}{0.880}
\newcommand{\PZeroRoneWallSaving}{0.711}

\newcommand{\PZeroQwenNullRecoveredWrong}{6}
\newcommand{\PZeroQwenActivatedAlreadyCorrect}{2}

\newcommand{\PZeroRoneNullRecoveredWrong}{5}

\newcommand{\ReadoutQwenFullCorrect}{84}

\newcommand{\ReadoutQwenFinalizeCorrect}{91}

\newcommand{\ReadoutRoneFullCorrect}{81}

\newcommand{\ReadoutRoneFinalizeCorrect}{87}

\newcommand{\FinalSeed}{1357911}
\newcommand{\FinalQwenWorkerCommandSeconds}{4884.56}

\newcommand{\FinalQwenFinalizeCorrect}{91}

\newcommand{\FinalQwenFinalizeActivated}{16}
\newcommand{\FinalQwenFinalizeSlots}{117718}

\newcommand{\FinalQwenFinalizeFullDeltaEstimate}{8.0}

\newcommand{\FinalQwenFinalizeFullGains}{8}

\newcommand{\FinalQwenSolutionFinalizeSlotSavingEstimate}{2.165}

\newcommand{\FinalQwenSlotsInteractionEstimate}{-4.660}

\newcommand{\FinalRoneWorkerCommandSeconds}{3671.65}

\newcommand{\FinalRoneFinalizeCorrect}{87}

\newcommand{\FinalRoneFinalizeActivated}{12}
\newcommand{\FinalRoneFinalizeSlots}{122195}

\newcommand{\FinalRoneFinalizeFullDeltaEstimate}{7.0}

\newcommand{\FinalRoneFinalizeFullGains}{7}

\newcommand{\FinalRoneSolutionFinalizeSlotSavingEstimate}{14.226}

\newcommand{\FinalRoneSlotsInteractionEstimate}{-3.230}

\newcommand{\RepeatSeed}{20261013}
\newcommand{\RepeatQwenWorkerCommandSeconds}{14121.54}
\newcommand{\RepeatQwenCurrentStageSlots}{347377}
\newcommand{\RepeatQwenFullCorrect}{83}

\newcommand{\RepeatQwenFirstBoxCorrect}{83}

\newcommand{\RepeatQwenSolutionCorrect}{90}

\newcommand{\RepeatQwenSolutionFirstBoxDeltaEstimate}{7.0}
\newcommand{\RepeatQwenSolutionFirstBoxDeltaLow}{3.0}
\newcommand{\RepeatQwenSolutionFirstBoxDeltaHigh}{12.0}
\newcommand{\RepeatQwenSolutionFirstBoxSlotSavingEstimate}{0.603}
\newcommand{\RepeatQwenSolutionFirstBoxSlotSavingLow}{0.332}
\newcommand{\RepeatQwenSolutionFirstBoxSlotSavingHigh}{0.868}
\newcommand{\RepeatQwenSolutionFirstBoxWallSavingEstimate}{0.660}

\newcommand{\RepeatQwenSolutionFirstBoxGains}{7}

\newcommand{\RepeatRoneWorkerCommandSeconds}{10621.61}
\newcommand{\RepeatRoneCurrentStageSlots}{353241}
\newcommand{\RepeatRoneFullCorrect}{75}

\newcommand{\RepeatRoneFirstBoxCorrect}{75}

\newcommand{\RepeatRoneSolutionCorrect}{85}
\newcommand{\RepeatRoneSolutionNull}{1}

\newcommand{\RepeatRoneSolutionFirstBoxDeltaEstimate}{10.0}
\newcommand{\RepeatRoneSolutionFirstBoxDeltaLow}{5.0}
\newcommand{\RepeatRoneSolutionFirstBoxDeltaHigh}{16.0}
\newcommand{\RepeatRoneSolutionFirstBoxSlotSavingEstimate}{0.833}
\newcommand{\RepeatRoneSolutionFirstBoxSlotSavingLow}{0.539}
\newcommand{\RepeatRoneSolutionFirstBoxSlotSavingHigh}{1.113}
\newcommand{\RepeatRoneSolutionFirstBoxWallSavingEstimate}{0.900}

\newcommand{\RepeatRoneSolutionFirstBoxGains}{10}

\newcommand{\ReadoutRepeatQwenFullCorrect}{85}

\newcommand{\ReadoutRepeatQwenSolutionCorrect}{90}

\newcommand{\ReadoutRepeatQwenSolutionFullGains}{5}
\newcommand{\ReadoutRepeatQwenSolutionFullHarms}{0}
\newcommand{\ReadoutRepeatQwenSolutionFullDelta}{5}
\newcommand{\ReadoutRepeatRoneFullCorrect}{77}

\newcommand{\ReadoutRepeatRoneSolutionCorrect}{85}

\newcommand{\ReadoutRepeatRoneSolutionFullGains}{9}
\newcommand{\ReadoutRepeatRoneSolutionFullHarms}{1}
\newcommand{\ReadoutRepeatRoneSolutionFullDelta}{8}

\newcommand{\BudgetCap}{\PZeroCap}
\newcommand{\PreCueSlots}{1984}

\newcommand{\OpenReviewURL}{}

\newcommand{\OpenReviewSlot}{%
  \ifdefempty{\OpenReviewURL}{}{\\[3pt]{\normalfont\small\href{\OpenReviewURL}{OpenReview discussion}}}%
}

\title{{\fontsize{20}{24}\selectfont Answer Delivery Under a Shared Token Budget:\
An Audited Study of Stopping and Finalization}}
\author{
    Yupeng Chang$^{1}$ \quad Linyuan Gao$^{1}$ \quad Zhiyuan Li$^{1}$ \\
    Yuan Wu$^{1}$\footnotemark[1] \quad Yi Chang$^{1,2,3}$ \\[3pt]
    {\normalfont $^{1}$School of Artificial Intelligence, Jilin University} \\[1pt]
    {\normalfont $^{2}$Engineering Research Center of Knowledge-Driven Human-Machine Intelligence, MOE, China} \\[1pt]
    {\normalfont $^{3}$International Center of Future Science, Jilin University} \\[4pt]
    {\normalfont \{changyp23, lygao25, zhiyuanl24\}@mails.jlu.edu.cn,} \\[1pt]
    {\normalfont \{yuanwu, yichang\}@jlu.edu.cn}%
    \OpenReviewSlot
}

\hypersetup{
  pdftitle={Answer Delivery Under a Shared Token Budget: An Audited Study of Stopping and Finalization},
  pdfauthor={Yupeng Chang, Linyuan Gao, Zhiyuan Li, Yuan Wu, Yi Chang},
  pdfsubject={arXiv preprint}
}
\begin{document}
\maketitle
\footnotetext[1]{Corresponding author}
\begin{abstract}
An inference budget is useful only when the system delivers an answer within it. We study a fixed-budget policy combining first-valid-box stopping with a charged finalization cue. On \PZeroEvalN{} fresh, independently audited GSM8K test questions per model, it improves delivered endpoint accuracy over first-box stopping by \PZeroQwenAccuracyDelta{} percentage points for Qwen3-4B-Thinking-2507 and \PZeroRoneAccuracyDelta{} for DeepSeek-R1-Distill-Llama-8B. A prelocked repeat on the same questions gives \RepeatQwenSolutionFirstBoxDeltaEstimate{} and \RepeatRoneSolutionFirstBoxDeltaEstimate{} points; neither seed shows a gross harm relative to first-box stopping. The point estimates of incremental physical-slot savings are below one percent. A finalization-only ablation separates cue recovery from stopping, and a fixed scalar reader tests extraction as an alternative explanation. The cue is prior art; the contribution is a component and readout analysis with charged injections and verified paths. Results cover two checkpoints, one task, one prompt family, and two fixed seed conditions.

\end{abstract}
\section{Introduction}
A reasoning system must return a usable answer before its inference budget expires. A long completion can fail this requirement in two different ways: it may continue after an answer is already available, or it may spend its budget reasoning without producing a readable answer. Stopping targets the first case; finalization targets the second. Combining them is practical, but accuracy or token count alone does not show which component drives each effect.

Our question is whether a conditional finalization cue improves the answer actually delivered by first-box stopping when the cue and answer must fit inside the original output budget. We examine three quantities separately: answer availability, numerical correctness, and physical generation cost. Removing missing answers need not remove incorrect answers. Similarly, shortening a completion after a box appears need not preserve the value selected at native termination. These distinctions matter when a serving interface must choose a single endpoint rather than retrospectively search a completed trace.

Budget forcing in s1 appends an end-of-thinking delimiter and optional answer cue \citep{muennighoff2025s1}. The Coupling Tax studies answer crowding under shared output limits via split-budget generation \citep{nie2026coupling}. Answer-convergence stopping probes prefixes with forced answers \citep{liu2025answer}; an interface study separates completion timing from forced and generated readouts \citep{carlon2026shorter}; PUMA separates answer readiness from reasoning convergence \citep{min2026stop}. Related early-exit, budget-aware, overthinking, and no-thinking work provides complementary context \citep{yang2026dynamic,sun2510stop,han2025token,chen2025not,ma2025reasoning}. We study a live-cache implementation without claiming a new cue or convergence detector.

We physically generate native completion, first-box stopping, their combined finalization policy, and a prelocked finalization-only ablation on the same questions. Injected tokens count toward the common cap; branches reset their cache and sampler seed. An independent audit verifies complete coverage, identities, exact prefixes, and forward accounting before reference scoring. Each policy is physically executed, with measured costs for its actual generation path.

Our contribution is a bounded component and readout study with reproducible accounting. Conditional finalization recovers boxed numeric endpoints; stopping accounts for most R1 workload savings and can discard a later correct Qwen answer. Historical pilots are not pooled into the fresh cohort. Deep-thinking tokens and Think@$n$ \citep{chen2602think} concern internal measures; the present study tests delivered endpoints and physical costs.

\section{Delivery and Cost Protocol}
\label{sec:protocol}
\subsection{Policies and returned values}
All policies operate under a post-prompt cap $C=\BudgetCap$. \textsc{Full} continues until native EOS or the cap and reads the last syntactically completed \texttt{\string\boxed\{\ldots\}} occurrence. An invalid or absent last box returns null. \textsc{FirstBox} follows the same native decoding process but returns as soon as a completed box passes a fixed numeric grammar; without one, it falls back to the native EOS/cap endpoint. The selectors are part of each policy and are fixed before evaluation.

\textsc{Solution} follows \textsc{FirstBox}, with one conditional intervention. If generation reaches slot \PreCueSlots{} without a valid numeric box, it injects a newline, \texttt{</think>}, two newlines, and \texttt{Final answer: \string\boxed\{}. The cue occupies nine slots for each tested tokenizer, leaving 55 of the reserved 64 slots for sampled continuation. Generation uses the existing KV cache and stops at the first valid numeric box; native EOS/cap provides the fallback if none appears. The cue uses only the observed prefix, with no reference, future trajectory, or correctness signal. Native EOS before the trigger is respected.

\textsc{Finalize} uses the same cue condition and budget but continues to native EOS/cap, applying \textsc{Full}'s final-box selector. It isolates finalization without first-box stopping.

We score the returned value by exact rational equality to the reference; null counts as incorrect. A cap is a finish reason, not an automatic error: a valid value is scored even at the limit. Availability means a grammar-valid boxed numeric endpoint; null need not imply absence of a correct unboxed answer. All arms share a correctness criterion while retaining their operational endpoint selectors. Alternative readouts appear in Appendix~\ref{app:sensitivity}.

\subsection{Physical budget and matched paths}
Let $P$ be prompt length and $N$ the emitted post-prompt slots. The accounting boundary is
\begin{equation}
N=N_{\rm sampled}+N_{\rm injected},\qquad N\leq C.
\end{equation}
In this singleton implementation, producing $N$ slots takes one prefill and $N-1$ incremental forwards, consuming $P+N-1$ model input tokens. The final emitted token is not forwarded. Forced tokens occupy the cache path but do not call the sampler or consume sampling RNG draws. The intervention therefore changes the trajectory without purchasing a second answer budget.

Each question's paths share a prompt and reset seed. \textsc{FirstBox} must equal the native prefix or fallback; \textsc{Solution} must equal \textsc{FirstBox} when unexposed. \textsc{Finalize} must equal \textsc{Full} when unexposed and continue the actual \textsc{Solution} prefix when exposed. All arms share the cap and decoder, but realized compute can differ.

Request wall time includes generation, monitoring, durable log flush, and result output. Whole-process time additionally includes setup, loading, integrity checks, and exit. Component timings are nested within those totals and are not added again. Physical slots and input tokens are workload counts, not estimates of FLOPs, monetary cost, or batched throughput.

\subsection{Paired estimands and qualification}
For question $i$, let $Y_i^a$ indicate correctness under policy $a$. For \textsc{Solution} ($S$) versus \textsc{FirstBox} ($B$), define gains $G=\sum_i(1-Y_i^B)Y_i^S$ and harms $H=\sum_iY_i^B(1-Y_i^S)$. The paired accuracy difference is
\begin{equation}
\widehat{\Delta}_{S,B}=\frac{1}{n}\sum_i(Y_i^S-Y_i^B)=\frac{G-H}{n}.
\end{equation}
For cost $K$, reduction is $1-\sum_i K_i^S/\sum_i K_i^B$, a ratio of totals rather than a mean of per-question percentages. We resample paired questions 10,000 times with fixed seed 20261012 for descriptive percentile 95\% intervals. Models are analyzed separately; repeated arms do not add independent questions.

Before scoring, the audit requires successful worker and supervisor completion, exact planned IDs, sealed source and input maps, pre/post model-file receipts, GPU identities, and matching started/completed forward attempts. Prefix and cue checks must also pass. Reference answers remain in a separate sidecar until qualification. Implementation, parser, and audit details appear in Appendix~\ref{app:repro}.

\section{Experiments and Results}
\begin{table*}[t]
\centering
\begin{tabular}{@{}llrrrr@{}}
\toprule
Model & Policy & Correct / 100 & No value & Mean slots & Mean wall (s) \\
\midrule
Qwen & \textsc{Full} & 83 & 14 & 1179.80 & 47.64 \\
Qwen & \textsc{FirstBox} & 82 & 14 & 1158.97 & 46.81 \\
Qwen & \textsc{Solution} & \textbf{90} & 0 & 1151.69 & 46.44 \\
R1 & \textsc{Full} & 80 & 12 & 1225.07 & 36.15 \\
R1 & \textsc{FirstBox} & 80 & 12 & 1054.47 & 31.16 \\
R1 & \textsc{Solution} & \textbf{87} & 0 & 1048.12 & 30.94 \\
\bottomrule
\end{tabular}

\caption{All three physically generated policies on 100 fresh GSM8K test questions per model. Correctness uses the numeric value actually returned; \emph{No value} means the endpoint returned null. Mean post-prompt slots include injected cue tokens. Mean wall time includes generation, monitoring, flush, and result output, excluding shared process setup and offline audit. Bold marks highest correctness among shown arms.}
\label{tab:p0}
\end{table*}

\subsection{Cohort, controls, and baselines}
We use pinned Qwen3-4B-Thinking-2507 and DeepSeek-R1-Distill-Llama-8B checkpoints \citep{qwen3technicalreport,deepseekai2025deepseekr1incentivizingreasoningcapability} on GSM8K \citep{cobbe2021training}. The cohort contains \PZeroEvalN{} distinct test questions per model, selected by a fixed question-hash ordering before any output was inspected and excluding previously used questions. The models share the cohort but remain separate analyses. No learned policy parameters or fresh-test calibration are used; the cue, trigger, cap, prompt, and seed were fixed before the preceding development check.

Inference uses singleton BF16 SDPA, temperature $0.6$, top-$p$ $0.95$, and reset seed \PZeroPrimarySeed{}. Six three-arm order permutations cycle by question rank. All 600 primary, 200 later \textsc{Finalize}, and 600 repeat paths completed and passed independent auditing. Both added stages were locked before primary scoring and reuse the cohort; later ablation wall contrasts are stage-confounded. Runtime details appear in Appendix~\ref{app:runtime}.

The primary contrast is \textsc{Solution} versus \textsc{FirstBox}; native-completion and component contrasts are secondary. \textsc{Finalize} returns \FinalQwenFinalizeCorrect{} correct Qwen and \FinalRoneFinalizeCorrect{} correct R1 answers, versus \PZeroQwenSolutionCorrect{} and \PZeroRoneSolutionCorrect{} for \textsc{Solution}. Relative to \textsc{Full}, \textsc{Finalize} gains \FinalQwenFinalizeFullDeltaEstimate{} percentage points for Qwen and \FinalRoneFinalizeFullDeltaEstimate{} for R1, matching \textsc{Solution}'s gains over \textsc{FirstBox}. First-box stopping loses one otherwise correct Qwen answer while saving workload. Appendix~\ref{app:finalize} gives the paired four-policy analysis. Published convergence mechanisms are related work, not reproduced cost baselines.

\subsection{Accuracy and harm}
Table~\ref{tab:p0} reports all arms with a common correctness criterion. Relative to \textsc{FirstBox}, \textsc{Solution} gains \PZeroQwenAccuracyDelta{} percentage points for Qwen, with paired 95\% interval [\PZeroQwenAccuracyLow{}, \PZeroQwenAccuracyHigh{}], and \PZeroRoneAccuracyDelta{} for R1, with interval [\PZeroRoneAccuracyLow{}, \PZeroRoneAccuracyHigh{}]. The respective discordant pairs are \PZeroQwenGains{} gains and \PZeroQwenHarms{} harms, and \PZeroRoneGains{} gains and \PZeroRoneHarms{} harms. Reporting both counts prevents net accuracy from concealing baseline-correct answers lost by the intervention.

Zero observed harm is a finite-cohort observation. A one-sided binomial upper bound at $\alpha=0.025$ is \PZeroQwenHarmUCB\% for each model; Bonferroni correction covers the two model-wise harm statements under the binomial sampling assumption. It concerns harm relative to the comparator, not absolute answer error. Each interval resamples questions within its task, checkpoint, prompt, and seed condition.

At locked seed \RepeatSeed{}, \textsc{Solution} again gains \RepeatQwenSolutionFirstBoxDeltaEstimate{} points for Qwen (95\% [\RepeatQwenSolutionFirstBoxDeltaLow{}, \RepeatQwenSolutionFirstBoxDeltaHigh{}]) and \RepeatRoneSolutionFirstBoxDeltaEstimate{} for R1 ([\RepeatRoneSolutionFirstBoxDeltaLow{}, \RepeatRoneSolutionFirstBoxDeltaHigh{}]), with zero observed harms relative to \textsc{FirstBox}. R1 still returns \RepeatRoneSolutionNull{} null endpoint. This repeat adds no questions; Appendix~\ref{app:repeat} reports complete arms and costs.

The secondary comparison exposes an endpoint tradeoff. Qwen \textsc{Full} is correct on \PZeroQwenFullCorrect{} questions, compared with \PZeroQwenFirstboxCorrect{} for \textsc{FirstBox}; one native-correct answer is lost at the earlier endpoint. R1 remains at \PZeroRoneFullCorrect{} correct under both. Thus a first box cannot be assumed to equal the eventual answer even when the two paths share an exact prefix.

\subsection{Where the additional answers come from}
In P0, we perform a descriptive, post-hoc decomposition of audited question pairs; it adds no independent observations. Qwen's \textsc{FirstBox} arm returns no value on \PZeroQwenFirstboxNull{} questions, and R1's on \PZeroRoneFirstboxNull{}. \textsc{Solution} reduces both null counts to zero. All \PZeroQwenGains{} and \PZeroRoneGains{} correctness gains occur on those previously null endpoints. Wrong-valued baseline answers receive no cue and remain wrong. Appendix~\ref{app:endpoints} gives the complete transition counts.

Availability and correctness therefore separate sharply. Of the formerly null questions, \PZeroQwenNullRecoveredWrong{} Qwen and \PZeroRoneNullRecoveredWrong{} R1 deliveries become readable but remain wrong. Converting null to a number is useful for an interface, yet it would overstate task improvement to count every such conversion as a recovered solution. The aggregate accuracy increase is specifically the correct subset of these conversions.

Cue activation occurs on \PZeroQwenSolutionActivated{} Qwen and \PZeroRoneSolutionActivated{} R1 paths, injecting \PZeroQwenSolutionInjected{} and \PZeroRoneSolutionInjected{} slots. Activation is not identical to a null baseline endpoint: Qwen has \PZeroQwenActivatedAlreadyCorrect{} activated paths whose no-cue answers are eventually correct. They remain correct after intervention. Nonactivated \textsc{Solution} paths match \textsc{FirstBox} in emitted IDs, text, and returned values. These checks localize the observed accuracy changes to intervention-exposed paths, while leaving open whether the cue surfaces existing partial computation or changes subsequent reasoning.

\subsection{Cost depends on the comparator}
\begin{table}[t]
\centering
\begin{tabular}{@{}llrr@{}}
\toprule
Model & Pair & Slots $\downarrow$ (\%) & Wall $\downarrow$ (\%) \\
\midrule
Qwen & B/F & 1.77 & 1.73 \\
Qwen & S/F & 2.38 & 2.51 \\
Qwen & S/B & 0.63 & 0.80 \\
R1 & B/F & 13.93 & 13.79 \\
R1 & S/F & 14.44 & 14.41 \\
R1 & S/B & 0.60 & 0.71 \\
\bottomrule
\end{tabular}

\caption{Reductions in total physical slots and measured request wall time. F=\textsc{Full}, B=\textsc{FirstBox}, S=\textsc{Solution}; $X/Y$ compares policy $X$ with baseline $Y$. These are singleton instrumented costs on the same paired questions, not serving-throughput estimates.}
\label{tab:cost}
\end{table}
Against \textsc{FirstBox}, \textsc{Solution} reduces slots by \PZeroQwenSlotSaving\% for Qwen (95\% interval [\PZeroQwenSlotLow{}, \PZeroQwenSlotHigh{}]) and \PZeroRoneSlotSaving\% for R1 ([\PZeroRoneSlotLow{}, \PZeroRoneSlotHigh{}]). Wall reductions are \PZeroQwenWallSaving\% and \PZeroRoneWallSaving\%. The incremental cue benefit is consequently accuracy recovery with only a small average workload change, not a large speedup.

Table~\ref{tab:cost} shows why a single headline efficiency number would mislead. R1 \textsc{FirstBox} already saves \PZeroRoneFirstVsFullSlotSaving\% of slots against \textsc{Full}, and \textsc{Solution} saves \PZeroRoneSolVsFullSlotSaving\%. Qwen's corresponding reductions are \PZeroQwenFirstVsFullSlotSaving\% and \PZeroQwenSolVsFullSlotSaving\%, with the one lost native-correct answer shared by both stopped arms. Most R1 savings arise from removing its native tail, whereas the cue contributes the new correct deliveries. Table~\ref{tab:physical} in the appendix records component totals for auditing.

\section{Discussion}
The experiment separates shortening an available-answer tail from recovering a missing delivery. Tail savings require endpoint preservation; recovery requires correctness on converted nulls. At the primary seed, R1 cost savings mostly come from the former and accuracy gains entirely from the latter. Qwen shows that earlier endpoints can lose a correct eventual answer.

At the primary seed, finalization-only reproduces the cue's accuracy gains without stopping, while retaining Qwen's later correct answer. Zero question-wise accuracy interaction describes this cohort; it is not a population-equivalence certificate. Accuracy recovery and tail savings remain separate objectives.

Formatting recovery remains a strong alternative to changed reasoning. A fixed post-hoc reader, preferring the last valid box and otherwise the last scalar literal, raises \textsc{Full} correctness to \ReadoutQwenFullCorrect{}/\ReadoutRoneFullCorrect{} (Qwen/R1) at the primary seed and \ReadoutRepeatQwenFullCorrect{}/\ReadoutRepeatRoneFullCorrect{} at the repeat, below actual \textsc{Solution} deliveries. However, \textsc{Solution} has one harm relative to this alternative reader for primary-seed Qwen and repeat-seed R1 (Appendix~\ref{app:cheap-reader}). This heuristic does not explain all gains; its failure cannot identify improved reasoning. The cue changes the live cache; a mechanism claim needs controls beyond extraction.

The prior-work boundary is narrow. s1 establishes forced budget termination; Coupling Tax studies answer crowding; answer-convergence and PUMA use answer agreement or reasoning-convergence signals. Our measurement documents what this combination returns, where gains occur, and which component accounts for costs. It does not rank those methods or establish a new general convergence mechanism.

For deployment, the cap and returned-value rule are interface choices. Recovering a number does not ensure correctness, and native continuation may preserve later corrections. Transfer requires rechecking grammar, prompt, and workload. Sub-percent wall differences against \textsc{FirstBox} need separate batched serving measurements before a latency claim.

\section{Conclusion}
Conditional finalization improves first-box endpoint accuracy by \PZeroQwenAccuracyDelta{} and \PZeroRoneAccuracyDelta{} points for Qwen and R1 at the primary seed; a locked repeat retains the direction. Point estimates of incremental slot savings remain below 1\%; larger R1 savings against native completion come mainly from stopping. These results measure delivered answers and physical costs, not reasoning convergence.

\label{sec:body-end}
\section*{Limitations}
\label{sec:limitations}
The primary P0 results concern 100 distinct GSM8K test questions per model, one prompt family, one primary seed per question, and two checkpoints. The additional fixed seed reuses these questions. Models, seeds, and repeated arms are not pooled into a larger independent sample. Question-bootstrap intervals are descriptive and do not provide unconditional guarantees for arbitrary tasks, prompts, models, or seeds. Zero observed primary gross harms do not imply zero population risk. Under a binomial sampling assumption, the original two-model Bonferroni procedure gives a 95\% family-wise one-sided upper bound of \PZeroQwenHarmUCB\% for each model; it is neither a distribution-free certificate for this deterministically selected cohort nor a joint bound over both seeds.

At the primary seed, only two cue-exposed Qwen baselines were correct; all 12 exposed R1 baselines were null. Unexposed paths are identical by construction. The aggregate bound targets harm over questions, not harm conditional on cue exposure or baseline correctness. Zero aggregate harm therefore gives limited evidence about conditional cue safety. The post-hoc scalar reader is a deliberately limited heuristic; its failure neither rules out better extraction nor identifies a latent reasoning improvement.

The Solution policy is a controlled endpoint intervention. It does not establish reasoning convergence, causal usefulness of deep-thinking-token measures, or a new algorithmic mechanism. The cue and related forced-finalization procedures are prior art. Relative to Full, Qwen has one original-seed gross harm; therefore the paper does not claim universal safety or harm-free operation. The study uses a single task, prompt family, two fixed seeds, and two checkpoints, and its singleton instrumented wall times do not generalize to batched serving or monetary cost.

The original development pilot and historical stopping study remain archived as engineering and contextual evidence; they are not pooled into the fresh P0 inferential sample. One locked seed repeat supports consistency in this setting, not generality across arbitrary seeds, tasks, or models. The repeat retains one R1 null, so the original elimination of missing boxed values is not universal. Finalization-only concerns the original seed; fixed scalar-readout sensitivity covers both seeds and retains harms against the alternative comparator.

The prelocked finalization-only ablation reuses the original questions and seed. Its accuracy and physical-slot contrasts are coupled, but it ran after the three primary arms; stage and hardware-load differences confound wall-time contrasts. Zero observed question-wise accuracy interaction does not establish population equivalence or exclude interactions in other prompts, seeds, or cue-exposed cohorts. The new wrapper's recorded command wall also excludes its orchestration checks outside that command and is not a full rental-cost measurement.

\section*{Ethical Considerations}
The study uses public mathematical word problems and released model checkpoints, without new human-subject data. Correct delivery means matching a benchmark numeric answer; it does not establish safe use in consequential applications. We retain failed and null deliveries and qualify complete evidence before scoring to reduce selective reporting. The experiments concern inference-time computation and do not train or modify model weights.

\section*{Acknowledgments}
OpenAI ChatGPT and Codex assisted with implementation and debugging of inference/audit code, local artifact preparation, literature search and primary-source citation checking, evidence organization, manuscript framing, and drafting/editing prose.
\bibliography{references}
\appendix
\section{Reproducibility Details}
\label{app:repro}

\subsection{Fixed Cohort and Analysis Scope}

The primary P0 cohort contains \PZeroEvalN{} distinct questions from the official GSM8K test parquet. The selector first verifies the pinned parquet and exclusion-source byte hashes, then reads only the question column for membership selection. It excludes the sealed prior-question hash set, deduplicates identical question text using the lowest source index, and sorts eligible questions by SHA256 of the UTF-8 string
\begin{quote}
\texttt{budget-finalization-confirm-}\allowbreak\texttt{20261012-v1|}\emph{questionSHA}.
\end{quote}
The first \PZeroEvalN{} entries form the fixed cohort. The full question SHA, source index, selection hash, request ID, and question text are retained. Selection precedes reading the answer column; references are written to a separate offline gold sidecar. The generator's exact nine-field request schema contains no reference, score, or prior output.

Both models use the same selected questions. Each model has three physically executed paths per question, with the tuple \texttt{(full, firstbox, solution)} permuted in the six orders produced by \texttt{itertools.permutations}; question rank modulo six selects the order. Each path resets its sampler and native cache. Models, arms, and repeated seeds are never counted as additional independent questions. There is no replacement question, outcome-dependent expansion, or automatic retry.

Historical development and stopping artifacts remain separate. P0 executes \textsc{Full}, \textsc{FirstBox}, and \textsc{Solution}. The later \textsc{Finalize} component ablation, locked before P0 quality scoring, reuses all original questions and seed and contributes no independent questions. The migrated engineering replay also adds no test questions.

\subsection{Checkpoints, Runtime, and Prompt}
\label{app:runtime}

The checkpoints are \texttt{Qwen/Qwen3-4B-Thinking-2507}, revision
\begin{quote}
\texttt{768f209d9ea81521153ed}\allowbreak\texttt{38c47d515654e938aea},
\end{quote}
and \texttt{deepseek-ai/}\allowbreak\texttt{DeepSeek-R1-Distill-Llama-8B}, revision
\begin{quote}
\texttt{6a6f4aa4197940add5772}\allowbreak\texttt{4a7707d069478df56b1}.
\end{quote}
Acquisition manifests pin all checkpoint and tokenizer filenames, sizes, and SHA256 values. The executed P0 runtime is Python 3.10.20, PyTorch 2.5.1+cu121, Transformers 4.51.3, and NVIDIA driver 550.54.14. Inventory receipts identify two NVIDIA GeForce RTX 4090 devices, with Qwen assigned physical GPU 2 and R1 physical GPU 3 by UUID. Initial and final ownership checks do not establish continuous device exclusivity.

Every P0 path uses BF16, SDPA attention, batch size one, temperature 0.6, top-$p$ 0.95, and seed \PZeroPrimarySeed{}. The sampler converts logits to FP32, divides by temperature, sorts them, and removes tokens after the nucleus threshold while retaining the token that crosses it. Sampling uses \texttt{torch.multinomial} with a separate CUDA generator initialized to the fixed seed for each path. CPU thread counts are one, tokenizer parallelism is disabled, and model loading is local and offline.

The exact user content is the original question followed by one newline and this suffix:
\begin{quote}
Reason step by step and give the final numeric answer in \texttt{\string\boxed\{...\}}.
\end{quote}
The template is supplied no system message or few-shot example. The tokenizer's pinned chat template receives one user message with \texttt{tokenize=True} and \texttt{add\_generation\_prompt=True}. Prompts are not truncated; the worker requires a nonempty encoding of at most 4096 tokens and archives its exact IDs. The alternative estimate-and-verify suffix retained in the development prompt specification is not used in P0.

\subsection{Cue and Native Execution}

\textsc{Solution} stops at the first completed numeric box. If no such box has completed after slot 1984 and generation remains live, it substitutes the inherited cue beginning at slot 1985. The cue's exact escaped representation is
\begin{verbatim}
\n</think>\n\nFinal answer: \boxed{
\end{verbatim}
where each \texttt{\string\n} denotes one line-feed character, rather than two literal characters. The inherited Qwen cue IDs are
\begin{quote}
\texttt{198, 151668, 271, 19357, 4226, 25,}\newline
\texttt{1124, 79075, 90},
\end{quote}
and the R1 cue IDs are
\begin{quote}
\texttt{198, 128014, 271, 19918, 4320, 25,}\newline
\texttt{1144, 80175, 90}.
\end{quote}
Each cue has nine tokens and occupies slots 1985--1993, leaving at most 55 sampled slots within the original 64-slot reservation. Cue IDs are copied from the qualified development preflight, checked against its seal, and required not to contain any model EOS ID. P0 does not re-encode or search for an alternative cue.

The cue is inserted during the same native decode call using its live KV cache, attention mask, and position sequence. Forced slots do not call the sampler and consume zero RNG draws. After each emitted token, the next incremental forward receives that token with the current cache, extends the mask by one, and advances the position ID. The last emitted token is not forwarded. There is no cache serialization, reload, hidden replay, or additional model call for monitoring. Generated text is decoded from generated IDs with special tokens skipped; physical token accounting still includes those IDs.

\textsc{FirstBox} must equal the exact first-box prefix of \textsc{Full}, or its EOS/cap fallback if no valid box appears. An activated \textsc{Solution} must equal \textsc{Full} through slot 1984 and contain the exact cue at the recorded positions. An unactivated \textsc{Solution} must equal \textsc{FirstBox}. These checks concern within-run pairing; changed runtime and hardware do not imply equality to historical sampled trajectories.

\subsection{Numeric Grammar and Endpoint Sensitivities}
\label{app:sensitivity}

The restricted parser accepts signed integers, finite decimal literals, scientific notation, and exact fractions written with one top-level slash or a balanced \texttt{\string\frac}, \texttt{\string\dfrac}, or \texttt{\string\tfrac} command. Fraction components are parsed recursively. It normalizes Unicode minus, removes \texttt{\string\left} and \texttt{\string\right}, permits one surrounding dollar-sign pair and enclosing balanced parentheses or braces, and accepts correctly grouped thousands separators. It rejects separate adjacent numbers, units, free-form arithmetic, empty values, malformed fractions, zero denominators, strings longer than 512 characters, and scientific exponents with absolute value above 1000. Parsing uses \texttt{Decimal} and \texttt{Fraction}, with exact rational equality and no tolerance or expression evaluation.

The online monitor reads generated text only and tracks balanced, unescaped \texttt{\string\boxed\{\ldots\}} groups. A policy stop commits when the first group closes with a valid numeric value. It returns that value at the actual emitted token boundary, which may also contain other text from the same token. If no policy stop occurs, the native EOS/cap endpoint selects the completed box with the latest opening position. An invalid final completed box returns null even if an earlier box was valid. An unfinished later box does not erase an earlier completed box under the primary delivered-value rule. No unboxed-number fallback is allowed.

Primary correctness compares the actual returned value to the reference regardless of finish reason. The supplemental last-valid-box readout instead ignores invalid completed boxes; native-EOS-only correctness additionally requires native EOS; the legacy nontruncated score rejects cap endpoints and any unfinished box. These are descriptive sensitivities, not replacements for the primary endpoint. The native-EOS restriction would reject intentional policy stops, so Table~\ref{tab:p0-endpoint-sensitivity} presents its informative comparison for \textsc{Full} only.

\begin{table}[t]
\centering
\begin{tabular}{@{}lrrrr@{}}
\toprule
Model & Primary & Valid & EOS & Legacy \\
\midrule
Qwen & 83 & 83 & 82 & 82 \\
R1 & 80 & 80 & 78 & 78 \\
\bottomrule
\end{tabular}

\caption{Descriptive endpoint sensitivities for \textsc{Full}. Primary is the delivered-value rule, Valid selects the last valid box, EOS requires native termination, and Legacy uses the nontruncated score. Entries count correct questions with the same \PZeroEvalN{}-question denominator.}
\label{tab:p0-endpoint-sensitivity}
\end{table}

\subsection{Qualification and Immutable Provenance}

Quality scoring begins only after both models' complete markers and successful supervisor exits exist. The independent audit requires exact complete-plan membership and order; acquisition, source, input, deployment, and output seals; recorded runtime and GPU bindings; prompt, seed, and question identities; complete started/completed forward ledgers; slot and model-input accounting; exact cue and paired-prefix coupling; and one answer emission per path. It independently reconstructs box events and numeric values from raw generated text, checks the actual stdout endpoint, verifies that answer publication precedes buffered evidence flushing, and requires no pending events after durable flush. The complete archive is checked again before crossing the gold-reading boundary. A qualification failure produces no partially scored table.

Mandatory CPU and native-cache/RNG tests and the complete label-free engineering replay precede fresh confirmation. The executing worker is an explicitly sealed compatibility copy: it resolves the science directory through a controlled environment variable, checks the actual amended Torch version, and preserves exception notes through Python 3.10's \texttt{\_\_notes\_\_} attribute. The separately recorded watchdog amendment changes the original 10800-second ceiling to 28800 seconds with 30 seconds for cleanup; the per-answer \PZeroCap{}-slot budget and per-model 614400-forward ceiling remain fixed. Original scientific cores, request membership, sampler, prompt, cue, and analysis rules are retained. Worker diffs and original/executing hashes distinguish these changes from the frozen scientific package.

The reproducibility archive separates frozen source and inputs, acquisition maps, executing deployment copies and diffs, qualification logs, and complete raw runs. The generated numerical interface records the independently qualified summary's SHA256, expected deployment/qualification/engineering/freeze/plan/request seals, bootstrap settings, and macro-to-source mapping. Thus manuscript values derive from the audited summary rather than hand-entered results. The offline archive validates the recorded acquisition and pre/post full-file maps; it does not independently rehash absent multi-gigabyte model bytes or re-execute tokenization.

\subsection{Costs and Statistical Recalculation}

Table~\ref{tab:physical} reports actual costs for all three executed arms. For a prompt of $P$ tokens and $N$ emitted slots, there are $N$ forwards: one prefill and $N-1$ incremental decode forwards. Physical model-input tokens total $P+N-1$. Slots include sampled, injected, and emitted EOS tokens. These counts are not estimates of FLOPs, energy, or monetary cost.

Answer-availability time ends after flushed stdout publication. Request wall includes the native wrapper, buffered evidence flush, and durable result output. Prefill, decode, sampling, text decoding, monitoring, and ledger component times are contained within these outer scopes and must not be added to them again. Whole-process supervisor wall additionally includes adapter imports and checks, checkpoint loading, all arms, post-run hashing, and exit: \PZeroQwenWholeSeconds{} seconds for Qwen and \PZeroRoneWholeSeconds{} seconds for R1. Its shared overhead cannot be attributed to one arm as a measured standalone deployment cost. Offline gold scoring and bootstrap computation occur after generation and are outside these request and worker timings.

\begin{table*}[t]
\centering
\begin{tabular}{@{}llrrrrr@{}}
\toprule
Model & Policy & Total slots & Sampled & Injected & Input tokens & Wall (s) \\
\midrule
Qwen & \textsc{Full} & 117980 & 117980 & 0 & 126946 & 4763.57 \\
Qwen & \textsc{FirstBox} & 115897 & 115897 & 0 & 124863 & 4681.28 \\
Qwen & \textsc{Solution} & 115169 & 115025 & 144 & 124135 & 4643.82 \\
R1 & \textsc{Full} & 122507 & 122507 & 0 & 130721 & 3614.71 \\
R1 & \textsc{FirstBox} & 105447 & 105447 & 0 & 113661 & 3116.08 \\
R1 & \textsc{Solution} & 104812 & 104704 & 108 & 113026 & 3093.92 \\
\bottomrule
\end{tabular}

\caption{Physical P0 costs from completed path and process receipts. Each arm contains \PZeroEvalN{} executed questions per model. Request wall includes generation, flush, and result output; the separately stated whole-process totals contain all three arms and their shared overhead. Nested timing scopes are not summed.}
\label{tab:physical}
\end{table*}

The bootstrap resamples paired questions within each model using \texttt{random.Random(20261012)} and 10000 resamples. The accuracy statistic is the mean policy-minus-baseline correctness difference. A cost reduction is one minus the ratio of the summed policy costs to summed baseline costs, both in the point estimate and each resample; it is not the mean of question-level ratios. The reported interval uses linearly interpolated 2.5th and 97.5th percentiles. These percentile intervals are descriptive.

Gross harm counts baseline-correct/policy-wrong questions, with all \PZeroEvalN{} questions as the denominator. The one-sided exact Clopper--Pearson bound uses $\alpha=0.025$ per model for the two-model family claim; $\alpha=0.05$ bounds are separate descriptive per-model quantities. Gains and harms remain separate, and exact two-sided McNemar tests use both discordant counts. The fixed primary comparison is \textsc{Solution} versus \textsc{FirstBox}; comparisons with \textsc{Full} remain secondary. Two fixed seeds and one task do not establish robustness across tasks, samplers, cues, reservation sizes, or checkpoints.

\subsection{Post-hoc Endpoint Decomposition}
\label{app:endpoints}

Table~\ref{tab:p0-endpoint-decomposition} describes transitions between the already qualified \textsc{FirstBox} and \textsc{Solution} endpoints. A wrong value is a present numeric value that differs from gold; a null is an absent returned value. Activation indicates that \textsc{Solution} inserted the cue, not that the baseline answer was wrong. A baseline may return a correct box after the cue boundary, so activation alone is not an accuracy gain. All wrong-valued baseline paths stop before cue exposure; their unchanged outcomes do not test the cue's ability to repair a wrong answer. Converting a null to a wrong value improves numeric-answer availability but remains incorrect under primary scoring. This post-hoc decomposition adds no independent question, subgroup confidence interval, hypothesis test, or out-of-sample effect. It uses audited endpoint exports and their recorded generated-ID/text hashes rather than a new token replay.

\begin{table*}[t]
\centering
\begin{tabular}{@{}llrrrr@{}}
\toprule
Model & FirstBox $\to$ Solution & $n$ & Active & Gain & Harm \\
\midrule
Qwen & Null $\to$ correct & 8 & 8 & 8 & 0 \\
 & Null $\to$ wrong value & 6 & 6 & 0 & 0 \\
 & Correct $\to$ correct & 82 & 2 & 0 & 0 \\
 & Wrong value $\to$ wrong value & 4 & 0 & 0 & 0 \\
\midrule
R1 & Null $\to$ correct & 7 & 7 & 7 & 0 \\
 & Null $\to$ wrong value & 5 & 5 & 0 & 0 \\
 & Correct $\to$ correct & 80 & 0 & 0 & 0 \\
 & Wrong value $\to$ wrong value & 8 & 0 & 0 & 0 \\
\bottomrule
\end{tabular}

\caption{Post-hoc descriptive \textsc{FirstBox}$\to$\textsc{Solution} endpoint transitions on the existing P0 cohort. $n$ counts questions, Active counts cue activations, and Gain/Harm count correctness discordances. Models remain separate; subgroup membership does not increase independent sample size.}
\label{tab:p0-endpoint-decomposition}
\end{table*}

\subsection{Prelocked Finalization-only Component}
\label{app:finalize}

\textsc{Finalize} (T) conditionally inserts the identical nine-token cue after slot 1984 but continues to EOS/cap and uses the primary final-box readout. Both checkpoints execute all 100 original questions at seed \FinalSeed{}, with the same prompt, sampler, checkpoint, cap, and native cache path. The source, inputs, threshold, budget, and test order were fixed before P0 scoring. The complete package passed 104 mandatory CPU tests before generation. A separate actual-run audit checks all 200 added paths and exact F/T identity when unexposed, pre-cue F/T identity when exposed, and the full S/T first-box prefix. A separately written raw-endpoint and question-bootstrap recalculation reproduces all counts and intervals. These are four policies on 100 paired questions per model, not 400 independent observations.

\begin{table*}[t]
\centering
\begin{tabular}{llrrrrr}
\toprule
Model & Policy & Correct & Null & Mean slots & Cue & Wall (s) \\
\midrule
Qwen & \textsc{Full} & 83 & 14 & 1179.80 & 0 & 47.64 \\
Qwen & \textsc{FirstBox} & 82 & 14 & 1158.97 & 0 & 46.81 \\
Qwen & \textsc{Finalize} & 91 & 0 & 1177.18 & 16 & 48.20 \\
Qwen & \textsc{Solution} & 90 & 0 & 1151.69 & 16 & 46.44 \\
R1 & \textsc{Full} & 80 & 12 & 1225.07 & 0 & 36.15 \\
R1 & \textsc{FirstBox} & 80 & 12 & 1054.47 & 0 & 31.16 \\
R1 & \textsc{Finalize} & 87 & 0 & 1221.95 & 12 & 35.62 \\
R1 & \textsc{Solution} & 87 & 0 & 1048.12 & 12 & 30.94 \\
\bottomrule
\end{tabular}

\caption{Complete same-seed component analysis. F/B/S are the inherited P0 paths; T is the 200-path later ablation. Mean slots include injected tokens, Cue counts activated questions, and wall is mean actual request time. Later T wall contrasts are confounded by stage; inherited P0 costs are not new costs.}
\label{tab:finalize}
\end{table*}

T versus F gains \FinalQwenFinalizeFullGains{} correct Qwen and \FinalRoneFinalizeFullGains{} correct R1 answers, with zero observed harms. S versus T loses one correct Qwen answer and has no R1 correctness discordance, while reducing physical slots by \FinalQwenSolutionFinalizeSlotSavingEstimate\% and \FinalRoneSolutionFinalizeSlotSavingEstimate\%. The lost Qwen answer is the same unexposed endpoint change already seen in B versus F. This separates observed cue recovery from shortening the native tail.

For each question, the prespecified accuracy interaction is $(Y^S-Y^B)-(Y^T-Y^F)$. Every observed value is zero; the empirical bootstrap therefore degenerates at zero. That resampling fact is not a population-equivalence certificate, especially with only \FinalQwenFinalizeActivated{}/\FinalRoneFinalizeActivated{} exposed questions. Absolute slot interactions average \FinalQwenSlotsInteractionEstimate{} and \FinalRoneSlotsInteractionEstimate{} slots per question. Ratios with different baseline denominators are not subtracted to construct an interaction. The later T request wall also changes relative to F despite little slot change; stage differences prevent interpreting that timing difference as an algorithmic speedup.

Current-stage physical forwards total \FinalQwenFinalizeSlots{} for Qwen and \FinalRoneFinalizeSlots{} for R1. Their recorded bounded worker-command supervisor walls are \FinalQwenWorkerCommandSeconds{} and \FinalRoneWorkerCommandSeconds{} seconds; these contain worker generation, setup and hashing but exclude pre-command/post-command orchestration checks. They must not be relabeled full orchestration or rental cost, or summed with nested request phases.

\subsection{Fixed Post-hoc Scalar Readout}
\label{app:cheap-reader}

To test a cheap extraction alternative, we freeze R-v1 before inspecting its quality. It applies the same reader to every complete audited F/B/S trace at both fixed seeds; after T's complete qualification, the unchanged reader also processes all original-seed T traces. This adds no model call, continuation, or independent question. The reader returns the last valid completed box; when none exists, it returns the last boundary-qualified signed scalar literal in the generated text. Scalars permit decimals, scientific notation, and grouped thousands under the same exact rational parser. Attached word identifiers, slash-adjacent numbers, malformed comma grouping, and multi-dot chains are rejected. Unboxed expressions and fractions are not evaluated; incidental numbers in prose may be selected. The full regular expression and protocol are in the reproducibility materials.

Extraction consumes a six-field label-free adapter containing model, arm, question identity, generated text, original returned value, and actual slot count. Gold is opened separately after complete extraction. The protocol, reader source, adapter, and extracted spans are hashed; eleven synthetic reader tests and an independent root recalculation passed. No parser, span, or output is selected by correctness. Original primary scores remain unchanged.

\begin{table}[t]
\centering
\begin{tabular}{llrrrr}
\toprule
Model & Arm & P & R & Null P & Null R \\
\midrule
Qwen & Full & 83 & 84 & 14 & 0 \\
Qwen & FirstBox & 82 & 83 & 14 & 0 \\
Qwen & Finalize & 91 & 91 & 0 & 0 \\
Qwen & Solution & 90 & 90 & 0 & 0 \\
R1 & Full & 80 & 81 & 12 & 0 \\
R1 & FirstBox & 80 & 81 & 12 & 0 \\
R1 & Finalize & 87 & 87 & 0 & 0 \\
R1 & Solution & 87 & 87 & 0 & 0 \\
\bottomrule
\end{tabular}

\caption{Fixed post-hoc R-v1 sensitivity on the existing 100-question cohort per model. P denotes primary delivered correctness, R the alternative reader, and Null the number of absent numeric values. This is a complete offline sensitivity, not a separately deployed policy or additional test sample.}
\label{tab:cheap-reader}
\end{table}

At the primary seed, R-v1 removes every null but adds only one correct answer per model in F/B (Table~\ref{tab:cheap-reader}). Its T scores remain \ReadoutQwenFinalizeCorrect{}/\ReadoutRoneFinalizeCorrect{}, identical to the primary T scores; all earlier F/B/S extracted spans are unchanged in the four-arm replay. Primary \textsc{Solution} versus R(\textsc{Full}) has seven gains and one harm for Qwen, and six gains and zero harms for R1; both net differences are six percentage points. Uniform R(\textsc{Solution}) versus R(\textsc{FirstBox}) differs by seven and six points. These comparisons are descriptive, post-hoc, and use the same original questions. R(\textsc{Full}) requires the full actually generated trace and inherits its model workload. Reader CPU time is separate from historical inference wall time, which cannot be presented as an independently measured deployed-reader latency.

\subsection{Prelocked Seed Repeat}
\label{app:repeat}

The added seed \RepeatSeed{} executes F/B/S on all original 100 questions per model, with the same sealed checkpoint, prompt, sampler, cue, threshold, cap, endpoint selectors, and six-order cycle. It was fixed before P0 scoring; no choice uses the observed accuracy. Within the new seed, B/F and S/B or pre-cue S/F must have exact emitted-ID/text coupling. Across seeds, question, prompt and model identities match, without requiring sampled trajectories to match. Both complete 300-path plans, successful bounded exits, 104 qualification tests, model pre/post file maps, UUID bindings, every physical attempt and cost, and actual published text/IDs/values pass independent qualification. A label-free guard rechecks all manifested bytes immediately before the original gold reader. A separately written raw-text endpoint and question-bootstrap calculation reproduces all counts, discordances, cost ratios, intervals and exact binomial quantities. These 600 new generation paths add no independent questions.

\begin{table*}[t]
\centering
\begin{tabular}{llrrrrr}
\toprule
Model & Policy & Correct & Null & Mean slots & Cue & Wall (s) \\
\midrule
Qwen & \textsc{Full} & 83 & 14 & 1181.86 & 0 & 47.81 \\
Qwen & \textsc{FirstBox} & 83 & 14 & 1149.42 & 0 & 46.55 \\
Qwen & \textsc{Solution} & 90 & 0 & 1142.49 & 17 & 46.24 \\
R1 & \textsc{Full} & 75 & 18 & 1290.80 & 0 & 38.45 \\
R1 & \textsc{FirstBox} & 75 & 18 & 1125.49 & 0 & 33.48 \\
R1 & \textsc{Solution} & 85 & 1 & 1116.12 & 18 & 33.18 \\
\bottomrule
\end{tabular}

\caption{Complete fixed-seed repeat on the same 100 questions per model. Slots include sampled, forced, and emitted EOS tokens. Wall is mean actual request wall; it does not include shared loading or the offline audit. This is a separate secondary seed condition, not another 100-question cohort or a replicated T arm.}
\label{tab:repeat}
\end{table*}

Qwen F/B/S correctness is \RepeatQwenFullCorrect{}/\RepeatQwenFirstBoxCorrect{}/\RepeatQwenSolutionCorrect{}, and R1 is \RepeatRoneFullCorrect{}/\RepeatRoneFirstBoxCorrect{}/\RepeatRoneSolutionCorrect{}. S versus B gains \RepeatQwenSolutionFirstBoxGains{}/\RepeatRoneSolutionFirstBoxGains{} answers with zero observed harms. The per-seed paired accuracy intervals are reported in the main text. Physical-slot savings are \RepeatQwenSolutionFirstBoxSlotSavingEstimate\% (95\% [\RepeatQwenSolutionFirstBoxSlotSavingLow{}, \RepeatQwenSolutionFirstBoxSlotSavingHigh{}]) and \RepeatRoneSolutionFirstBoxSlotSavingEstimate\% ([\RepeatRoneSolutionFirstBoxSlotSavingLow{}, \RepeatRoneSolutionFirstBoxSlotSavingHigh{}]); measured request-wall savings are \RepeatQwenSolutionFirstBoxWallSavingEstimate\% and \RepeatRoneSolutionFirstBoxWallSavingEstimate\%. B versus F has no correctness discordance in this seed, which does not erase the original Qwen loss or establish endpoint equivalence.

R1 retains one S null despite cue activation: it emits a completed mixed-number box containing a unit, outside the fixed numeric grammar, then reaches native EOS at 2008 slots. This is not missing cue exposure, lack of any visible output, or a reason to change the parser after evaluation. The original elimination of boxed nulls therefore does not generalize even to this second seed. The observed improvement direction is repeated; two fixed seeds do not establish arbitrary-seed robustness. The original primary comparison and intervals remain unchanged, and the two seeds or models are not pooled as independent samples. The fourth-arm component analysis concerns only the original seed. A further prelocked seed was not submitted because the runtime and audit budget would conflict with finalization; this scheduling decision preceded new quality inspection.

Current-repeat forwards total \RepeatQwenCurrentStageSlots{} for Qwen and \RepeatRoneCurrentStageSlots{} for R1. Recorded bounded worker-command walls are \RepeatQwenWorkerCommandSeconds{} and \RepeatRoneWorkerCommandSeconds{} seconds. As with T, these exclude outer orchestration checks and must not be added to nested request timings or relabeled full rental cost. Prior P0/T costs remain prior costs, not new repeat costs.

\subsection{Fixed Readout on the Seed Repeat}
\label{app:repeat-reader}

The unchanged R-v1 reader also processes all 600 complete, independently qualified new-seed F/B/S traces. An extension scope is sealed before extraction; every label-free span is independently checked before opening gold. The original reader, protocol, numeric grammar and primary scores remain unchanged. A separate recount verifies every model/arm count and paired correctness discordance. This is a post-hoc sensitivity on the same 100 question clusters per model, with no new generation, independent question, confidence interval, or measured deployed-reader wall time.

\begin{table}[t]
\centering
\begin{tabular}{llrrrr}
\toprule
Model & Arm & P & R & Null P & Null R \\
\midrule
Qwen & Full & 83 & 85 & 14 & 0 \\
Qwen & FirstBox & 83 & 85 & 14 & 0 \\
Qwen & Solution & 90 & 90 & 0 & 0 \\
R1 & Full & 75 & 77 & 18 & 0 \\
R1 & FirstBox & 75 & 77 & 18 & 0 \\
R1 & Solution & 85 & 85 & 1 & 0 \\
\bottomrule
\end{tabular}

\caption{Unchanged R-v1 readout at the second fixed seed. P, R and Null have the same meaning as Table~\ref{tab:cheap-reader}. Every row has 100 questions; adding seeds or policies does not increase independent question count.}
\label{tab:repeat-reader}
\end{table}

R(\textsc{Full}) and R(\textsc{FirstBox}) return \ReadoutRepeatQwenFullCorrect{} correct Qwen and \ReadoutRepeatRoneFullCorrect{} correct R1 answers, whereas R(\textsc{Solution}) retains \ReadoutRepeatQwenSolutionCorrect{}/\ReadoutRepeatRoneSolutionCorrect{}. Relative to R(\textsc{Full}), actual \textsc{Solution} has \ReadoutRepeatQwenSolutionFullGains{} gains and \ReadoutRepeatQwenSolutionFullHarms{} harms for Qwen, and \ReadoutRepeatRoneSolutionFullGains{} gains and \ReadoutRepeatRoneSolutionFullHarms{} harm for R1; the net differences are \ReadoutRepeatQwenSolutionFullDelta{} and \ReadoutRepeatRoneSolutionFullDelta{} points. Uniform R(\textsc{Solution}) versus R(\textsc{FirstBox}) gives the same discordances here. Thus zero primary-comparator harms cannot be transferred to an alternative reader. R-v1 removes all nulls, including the remaining R1 \textsc{Solution} null, but that last conversion is wrong. Availability without correctness is again insufficient. Only this fixed heuristic is tested; a stronger parser or extraction model could behave differently.

\end{document}